\documentclass{article} 
\usepackage{aaai21}  
\usepackage{times}  
\usepackage{helvet} 
\usepackage{courier}  
\usepackage[hyphens]{url}  
\usepackage{graphicx} 
\usepackage{natbib}  
\usepackage{caption} 
\usepackage{subcaption}
\usepackage{color}
\newcommand{\hk}[1]{{#1}} 
\usepackage{balance}
\usepackage[switch]{lineno}

\title{A Subjective Model of Human Decision Making Based on\\ Quantum Decision Theory}

\author{Chenda Zhang ~~~~ Hedvig Kjellström\\}

\affiliations{Division of Robotics, Perception and Learning\\
KTH Royal Institute of Technology, Sweden\\
{\tt chenda@kth.se} ~~~~ {\tt hedvig@kth.se}}

\begin{document}
\maketitle

\begin{abstract}
\hk{Computer modeling of human decision making is of large importance for, e.g., sustainable transport, urban development, and online recommendation systems. In this paper we present a model for predicting the behavior of an individual during a binary game under different amounts of risk, gain, and time pressure. The model is based on Quantum Decision Theory (QDT), which has been shown to enable modeling of the irrational and subjective aspects of the decision making, not accounted for by the classical Cumulative Prospect Theory (CPT). Experiments on two different datasets show that our QDT-based approach outperforms both a CPT-based approach and data driven approaches such as feed-forward neural networks and random forests.}


\end{abstract}

\section{Introduction}

\hk{Computer modeling of human decision-making has a long history, but is today of unprecedented importance with applications in areas such as sustainable transport, urban development, and online recommendation systems. }

\hk{The established methodology for modeling of decision making under risk is Cumulative Prospect Theory (CPT) \cite{kahneman2013prospect} in which value is a function of the gains and losses of available actions that the decision-maker could take. Stochastic versions of CPT, such as CPT with a logit function (logit-CPT) \cite{carbone}, account for the inherent randomness in human decision-making. The theory is coherent and comprehensive and has been shown to model a large class of human decision-making problems. 

However, there have been observations \cite{birnbaum2004tests} of systematic deviations from the outcome predicted by utility theories, e.g., interference between different variables in complex decision making. In other words, the observed decision-makers did not behave completely rationally.}

\hk{In response to this, the Quantum Decision Theory (QDT) \cite{yukalov2008quantum} was developed. QDT is summarized in Section 4 below and in the Appendix, but can shortly be expressed as:} 
Each option \hk{for the decision-maker} corresponds to a {\em prospect probability}, which is the sum of two factors: the {\em utility factor} representing the classical utility of the option, and the {\em attraction factor} \hk{(i.e., the core addition of QDT)} representing irrational and subjective aspects that affect the 
prospect attractiveness for the decision-maker. 

\hk{Earlier QDT-based prediction methods \cite{favre2016quantum,Quant,vincent2016calibration}}
%
%
used 
attraction factor functions that only captured one aspect of the decision maker's state of mind. \hk{The contribution of the present paper is a multi-modal attraction factor function that take multiple emotional and cognitive aspects into account,}
such as the framing effect \cite{tversky1981framing} or stress \hk{caused by a} 
time constraint. 
We propose four parameterized attraction factor components corresponding to different irrational effects \hk{in} 
the decision making process. \hk{This is further described in Section 5.}


\hk{Experiments, presented in Section 6, confirm that the inclusion of the proposed attraction factor enabled our QDT model to foresee the behavior of individual players to a higher degree than a CPT model without attraction factor. Our model also outperforms purely data driven methods such as feed-forward neural networks and random forests.
}

\section{Related Work}

\hk{Prospect Theory, with its development} Cumulative Prospect Theory (CPT) \cite{tversky1992advances}, \hk{can be regarded as the classical approach to modeling human decision making under risk. It features} 
several parameters that isolate and quantify psychological concepts such as loss aversion, subjective value functions for gains and losses, and probability weighting functions. 
\hk{These parameters can be estimated from data on an individual level} \cite{harless1994predictive,harrison2009expected} 
if there are enough data supported for each subject\hk{, or in a} 
hierarchical \cite{nilsson2011hierarchical,murphy2018hierarchical} \hk{fashion}
, first on the aggregate level and \hk{then refined for each individual. Mixture models have been used} 
\cite{harrison2009expected,conte2011mixture} \hk{to account for} 
the considerable heterogeneity observed in people's risk-taking behavior. 

Despite \hk{its successes in} 
modeling and predicting human decisions under uncertainty, Birnbaum (\citeyear{birnbaum2004tests}) concludes that CPT
, even with well-fitted parameters, is \hk{unable to account for a number of observed phenomena, such as interference between variables in complex decision making, the immediate regret caused by an unsuccessful decision  , or stress \hk{caused by a}
time constraint. This} 
implies that the prospect theory is intrinsically incomplete. 

In \hk{response to this}
, Yukalov and et al.~(\citeyear{yukalov2008quantum,yukalov2009processing}) introduced the Quantum Decision Theory (QDT) 
with a goal of providing a complete framework for modeling and predicting human decisions. Its use of Hilbert spaces constitutes a generalization of the probability theory axiomatized by Kolmogorov (\citeyear{kolmogorov2018foundations}) for real-valued probabilities to probabilities derived from algebraic complex number theory. By its mathematical structure, QDT aims at encompassing the superposition processes occurring down to the neuronal level. Numerous behavioral patterns, including the behavioral phenomena unexplained by utility-theoretic approaches, are coherently explained by QDT \cite{yukalov2016quantum}. \hk{A summary of QDT can be found in Section 4 of this paper.}
The authors 
demonstrated \hk{that} QDT has the potential to explain the results of \hk{certain} 
non-trivial decision problems 
\cite{shafffi1990typicality}, \hk{unaccounted for by traditional models}. 

Favre and et al.~\citeyear{favre2016quantum} applied \hk{QDT to a lottery game with} 
choices between a certain and risky lottery. 
The 
result was in good agreement with the QDT predictions at the level of groups. However, on the individual level, the 
authors \hk{pointed out the need to calibrate} 
the attraction factors and characterize individual decision-makers.

Yukalov and et al.~\cite{Quant} extended \hk{this work to games with both gain and loss, and refined the methodology for computing attraction and utility factors. The performance by QDT was} 
compared \hk{to} 
the famous experiments \hk{with CPT} done by Kahneman and Tversky \cite{kahneman2013prospect}. The results show that the predictions were in good agreement with the empirical data on an aggregated level. 

The first calibration and parameterization of QDT \hk{on an empirical dataset} was conducted in \cite{vincent2016calibration}. 
\hk{Their model combines logit-CPT to model the utility factor with a} 
constant absolute risk aversion (CARA) function \hk{to account for the attraction factor}. The CARA function represents the decision-makers' will to avoid large risks. 
\hk{Experiments on a binary lottery dataset showed the proposed QDT model to outperform a pure logit-CPT} 
model on both group and individual level. \hk{The difference was especially prominent for decision tasks with large risk,} 
thanks to the CARA attraction factor function. 

\hk{Vincent et al.~pointed out the need for the inclusion of further factors characterizing the state of mind of the decision maker, which will explain variations in the attraction factor not related to the structure of the decision problem. In the present paper, we proceed in this direction by including factors affecting the player's cognition, such as time pressure and memory effects.}

\section{Datasets}


\hk{We use data collected by \cite{diederich2020need} from two experiments, both involving games with two-choice options, a sure option (where there is a lower but quaranteed gain) and a gamble option (where there is a higher gain but with some uncertainty). The game was always fair, in that the sure option and the gamble option always had the same expected value. The experiments were also designed with time constraints and minimum scores. 

In addition, the sure option was presented either with a positive or negative connotation to it, introducing a framing effect \cite{tversky1981framing}. For example, in a game with an initial amount of 100 points and winning probability of 0.4, the sure option would be presented as "Keep 40" in the positive framing, and "Lose 60" in the negative.

The data contain details on the setup of every individual game trials and response from the participant.}


\subsubsection{Dataset 1.}
 \hk{19 participants were involved in the first experiment.} Each participant needed to take three game sessions, where each session contained four blocks of trials. 
 \hk{The participants were reimbursed per point gained in the games.} The games had four different point amounts (25, 50, 75 or 100) and four probabilities of winning (0.3, 0.4, 0.6, or 0.7).
 
 \hk{For each game trial, a sure and a gamble option were created as described above, with equal expected gain in the sure and gamble options and sure option presented either in a gain or loss context. 
 The experiment designers also designed eight catch trials with non-equivalent sure and gamble options, to validate the systematicity of the decision-making. There were thus in total $4*4*2+8=40$ different trials. Within each block, each game was presented twice in a random order, resulting in 80 trials per experimental block. 
 
 Moreover, two different response time constraints (1 s or 3 s) and three levels of need (0, 2500, or 3500) were induced, defined as the minimum number of points needed \hk{in order to keep the earned points} during the \hk{games in this} block. The blocks of trials were repeated twice for each response time constraint and once for each need level, yielding 
 $80*4*3=960$ game trials 
and in total $19*960=18240$ data points.}
 
\subsubsection{Dataset 2.}
\hk{Experiment 2 used the same design, but with 58 (new) participants. Each trial had initial amount 19, 20, 21, 39, 40, 41, 59, 60, 61, 79, 80 or 81, and winning probability 0.3,  0.4, 0.6, or 0.7, forming $12*4*2=96$ different game setups with either positive or negative framing. Each block containing 96 regular game trials (with each setup present once) and 8 catch trials.

There were two time limits (1 s or 3 s) and three need levels (0, 2800, or 3600 points). Blocks were repeated for each unique combination, yielding six blocks with a total of $(96+8)*6=624$ observations per participant. In total, there were thus $624*58=36192$ data points.}

\section{Quantum Decision Theory}
Quantum Decision Theory (QDT) \cite{yukalov2008quantum} is an intrinsically probabilistic framework that is derived from treating prospects and decision-makers' state of mind as vectors in a complex Hilbert space. A more detailed explanation and mathematical derivation will is given in Appendix, while we are trying to give a brief overview of the theory in this section. When facing a decision problem, the decision-maker is assumed to be in a decision-maker state $\Phi$, which is a superposition of prospect states $\pi_i$, each prospect state corresponds to an option in the decision problem. When the decision-maker makes a decision, $\Phi$ collapse into one of the prospect states, which means he/she chooses the option corresponds to that prospect state. 

A general formula for $\pi_i$'s probability under QDT is:
\begin{equation}
    P(\pi_i)=f(\pi_i)+q(\pi_i)
\end{equation}
"f" is the utility factor, and "q" is the attraction factor. There are also some constraints applying to the factors within the QDT framework.

Since $P(\pi_i)$ are probabilities, they must satisfy:
\begin{equation}
\sum_{i} P(\pi_i)=1, 0\leq P(\pi_i)\leq 1
\end{equation}
As \cite{yukalov2009processing} mentioned, the quantum decision theory reduces to the classical utility theory when the attraction factors vanish, thus the utility factors play the role of classical probability and must satisfy:
\begin{equation}
\sum_{i} f(\pi_i)=1, 0\leq f(\pi_i)\leq 1
\end{equation}
The attraction factors characterizes the attractiveness of the prospect, which is based on irrational subconscious factors. They follow the alternation law \cite{Quant}:
\begin{equation}
\sum_{i} q(\pi_i)=0, -1 \leq q(\pi_i)\leq 1
\end{equation}
We will now introduce the formulas we used for the utility factor and attraction factor.

\subsubsection{Utility factors.}
    
    
{We use the utility function from Prospect Theory \cite{kahneman2013prospect}. 
The utility function $U_A$ of a option A has the following formula in \hk{the} gain only or loss only domain:
 \begin{equation}
    U_{A}=w\left(p_{1}^{A}\right) v\left(V_{1}^{A}\right)+\left(1-w\left(p_{1}^{A}\right)\right) v\left(V_{2}^{A}\right)
    \end{equation}

    The value function $v(x)$ reflects how people value the gains and losses according to a reference points(often set to 0).  While the weighting function w(p) reflects people's altitude towards different probabilities, such as overestimate very small probabilities. There are many different variations of the value functions and the weighting functions. One good combination is a power value function combined with a Prelec II weighting function. The value function v(x) has the following formula:
$$v(x)=\left\{\begin{array}{rll}
x^{\alpha} & x \geq 0 & \alpha>0 \\
-\lambda(-x)^{\alpha} & x<0 & \lambda>0
\end{array}\right.$$
    The weighting function is:
    \begin{equation}
    w(p)=\exp \left(-\delta(-\ln (p))^{\gamma}\right), \quad \delta>0 \quad \gamma>0
    \end{equation}

    We propose that the decision maker does not always value the utility in the same way;there are chances that the option with lower utility function value appear to have higher utility in the decision maker's perspective. Therefore we supplemented the logistic function as a stochastic choice function. It mimics the partition function from statistical mechanics. The option with higher utility function value will have a higher f value:
    \begin{equation}
f_A=\frac{1}{1+e^{\varphi}\left(U_{B}-U_{A}\right)}
\end{equation}
    The f value for the other option becomes:
    \begin{equation}
    f_B=1-f_A
    \end{equation}
    
    There are four parameters that can be fitted in the utility factor: $\alpha,\delta,\gamma,\phi$.}
    As the utility factor in our QDT model, We set the reference point at 0, so all the options will only be considered in the gain domain. Because we want the utility term to only capture the utility of the option and no option will lead to any point loss, in experiment definition. Note that we are not using CPT as the utility term as we are only using it in the gain domain. However, in the baseline CPT model, we set the reference point at the expected value of each gamble, thus two possibilities of the gamble option fall into different domains where the value function behaves differently.

\subsubsection{Attraction factors.}

Following the derivation in Yukalov's work in 2016 \cite{yukalov2016quantum}, the attraction factor has the following constraints:
    \begin{equation}
    q_A=min(f_A,f_B)cos(\Delta^A), q_A+q_B=0
    \end{equation}
    where $cos(\Delta^A)$ represents the the argument of the uncertainty parts of the event in the Hilbert Space. When there are only two options, $cos(\Delta^A)=-cos(\Delta^B)$. The exact formula for $\Delta^A$ is impossible to be determined as the irrational part of decision making is highly unstable and complex. 
    
    However, we could approximate it with the help of our prior knowledge of the decision problem, in addition to some observations of human decision behaviors. We propose that the attraction factor could be estimated by assembling different attraction components, where each component represents one subconscious factor that affects the decision-maker during the decision-making process. Theoretically, if we could address all such subconscious factors and model them with exact formulas, then the $cos(\Delta^A)$ can be approximated perfectly.  However, human minds are highly complex, and it is not realistic to find exact formulas that correctly describe them. One practical method is to define parameterized formulas for each attraction factor component and fit the parameters for each person. 
    
We have addressed the primary subconscious factors in our decision problem and designed parameterized formulas. 

We will introduce some notations first: $\pi_A$: The prospect state corresponds to the gamble option,$\pi_B$:The prospect state corresponds to the sure option, STD: The standard deviation of the option, the sure option would have 0 in this term,
    $I_{framing}$=1 if the sure option is presented in a gain frame, =-1 when lose frame is presented. TL: the time limit set on making the decision, in our experiment, this is either 1 or 3. $I_{previous}=1$ if the player won the last gamble, equals -1 if he/she lost. When the player chose the sure option in the last game trial, $I_{precious}=0$. $S_{need}:$ the difference between the current score and the minimum needed score. $S_{init}$: The initial amount of each game trial.
    \begin{itemize}
        \item \textbf{The framing effect component.} Framing effect is a cognitive bias that the decision-maker changes their opinion on an option based on whether the option is presented in a positive or negative frame.\cite{tversky1981framing} For example, losing 200 dollars in a total of 1000 dollars could be described as "Keep 800 dollars" in a gain frame or "Lose 200 dollars" in a losing frame. Experiments show that people tend to avoid risk when a gain frame is presented while they tend to seek risk when a lost frame is presented. In the experiments we study, the sure option has been presented in either gaining frame or a losing frame. Thus the framing effect is an important subconscious factor in the decision-making process. We model this component with the following formula:
        
        \begin{equation}
        q_{framing}=-1\cdot I_{framing}\cdot STD^{c_1}, c_1\geq 0
        \end{equation}
        
        In a losing frame, $q_{framing}$ becomes greater when the uncertainty is larger and vice versa, reflecting people's risk-seeking behavior when a losing frame is presented. On the other hand, in a gain frame, $q_{framing}$ becomes greater when the risk in the option is small.
        \item \textbf{The time pressure component.} Time constraint plays an important role during a decision making process. People appeared to be less rational when they are making decisions under an extreme time pressure.\cite{svenson1993time} The shorter the time constraint is, the greater the pressure affecting the decision-maker. \cite{diederich2020need} observe that the framing effect is stronger under a shorter time limit, while the time limit itself does not affect the decision strongly. Therefore we model the time pressure component as an amplifier to the framing effect component, with the following form:
        
        \begin{equation}
            q_{time}=Exp(-c_2\cdot TL),c_2\geq 0
        \end{equation}

        \item \textbf{The memory effect component.} One of the most difficult parts of modeling human decision making is the involvement of memory. Different people have entirely different memories, thus have different feelings on the same option. In our experiment, the strongest memory would be the result of the previous game trial. The participant is more willing to choose a sure option after a losing game trial to avoid consecutively lost while seeking more risk after winning a game to maintain the luck. The memory effect component is modeled as:
        \begin{equation}
            q_{memory}=c_3\cdot I_{previous}\cdot STD
        \end{equation}
        One important property of this component is that it also incorporates the uncertainty of the gamble into the formula, where the decision maker would perceive the uncertainty in different ways based on the outcome of the previous game trial.
        
        \item \textbf{The need component.} In both experiment 1 and experiment 2, participants need to reach the minimum needed scores for getting the bonus reward. It is obvious that this need criterion would be an important subconscious factor in the decision-making process. When there is still a large gap between the current score and the target score, the decision-maker would be more inclined to choose the gamble option, and vice versa when the gap is small as choosing the sure option would be enough to reach the minimum needed score. The formula for this component is:
        \begin{equation}
            q_{need}=c_4\cdot (S_{need}-5\cdot STD\cdot (1-P_{gamble} )
        \end{equation}
        The $-5\cdot STD\cdot (1-P_{gamble}$ term accounts people's feeling of expected incoming points, which is subtracted from the real gap between current points and expected one.
        
    \end{itemize}

    
    When combining the components, we assume an additive relationship between $q_{frame},q_{memory}$ and $q_{need}$. The value of the $cos(\Delta^A)$ is between -1 and 1, thus we need a normalization function. In addition, $cos(\Delta^A)$ represents the relative difference between two options in the mind space, thus we are only interested in the difference between attraction factor values. The approximation for $cos(\Delta^A)$ is then
    \begin{equation}
        cos(\Delta^A)=tanh(a\cdot (q_{total}^{A}-q_{total}^{B}))
    \end{equation}
    \begin{equation}
        q_{total}=q_{time}\cdot q_{frame}+q_{memory}+q_{need}
    \end{equation}

\section{Methods}

Next, we present the various methods employed.
\footnote{\bf The source codes will be available upon request }

\subsubsection{Data processing.}
The raw experiment data from \cite{diederich2020need} was in Microsoft Excel format, where each row contains the complete information about one game trial, including the initial amounts, probability of winning the gamble, time limit and the minimum needed score in the game block, etc. We also removed three subject's data from Dataset 2 because of missing data entries. Therefore there are only 55 subjects in Dataset 2. We then process the data, for both Datasets 1 and 2, with the following steps:
\begin{enumerate}
    \item Group the data points according to the subject ID.
    \item Calculate new attributes for each row, e.g., the gap between the current score and the required minimum score or an entry that tells the result of the previous game trial, which is used to calculate the memory effect component.
    \item Shuffle the dataset and split into six data blocks. We used k-fold cross-validation to evaluate our model. Each data block will be used as a test dataset once while the other five are used for training.
\end{enumerate}

\begin{table*}[t]
    \centering
\caption{Accuracy comparison for different QDT models}
 \begin{subfigure}[t]{0.48\textwidth}
    \centering
 \subcaption{Dataset 1}
    \begin{tabular}{|c|c|}
    \hline
    Attraction factor & Accuracy \\
    \hline
    None & $0.757\pm 0.009$\\
    Time+Frame& $0.802\pm 0.002$\\
    Memory&$0.786\pm 0.004$\\
    Need& $0.783\pm 0.002$\\
    Time+Frame+Memory&$0.802\pm 0.004$\\
    Time+Frame+Memory+Need&$0.812\pm 0.008$\\
    \hline
    \end{tabular}
\label{accuracy_data1}
\end{subfigure} ~~~~~~
\begin{subfigure}[t]{0.48\textwidth}
    \centering
 \subcaption{Dataset 2}
    \centering
    \begin{tabular}{|c|c|}
    \hline
    Attraction factor & Accuracy \\
        \hline
    None&$0.677\pm 0.005$\\
    Time+Frame& $0.717\pm 0.017$\\
    Memory&$0.688\pm 0.002$\\
    Need& $0.691\pm 0.006$\\
    Time+Frame+Memory&$0.719\pm 0.006$\\
    Time+Frame+Memory+Need&$0.724\pm 0.006$\\
    \hline
    \end{tabular}
\label{accuracy_data2}
\end{subfigure}
\end{table*}

\subsubsection{Estimation of parameters.}
{Our estimation method and notations are inspired from \cite{vincent2016calibration}. The response at each game trial is denoted as $\Phi^{i}_j$. $\Phi^i_j=1$ if the player i chooses the gamble option in game trial j, and equals 0 if the sure option is chosen in that game trial.
We used a maximum likelihood method for estimating the parameters. The target function used was:
\begin{equation}
    \Pi^{q}_i=\Pi_{j} P_{ij}(\pi_A)^{\Phi^i_j}P_{ij}(\pi_b)^{1-\Phi^i_j}
\end{equation}

This estimation is done for every participant. As our goal is evaluating different attraction factor components, the attraction parameters needed to be estimated are also different. For example, if the attraction factor only consists of the memory effect component and the need component, only $c_4$ and $c_3$ need to be estimated. However, the four utility parameters are always estimated.

{\em Regularization.~~}
One major difficulty for the parameter estimation at the individual level was over-fitting. Since the amount of data for each individual subject is limited, one response could greatly impact the value of the estimation. To address this problem, we added a regularization term in the target function, so it becomes

\begin{equation}
    \Pi^{q}_i=-log(\Pi_{j} P_{ij}(\pi_A)^{\Phi^i_j}P_{ij}(\pi_b)^{1-\Phi^i_j})+\sum_{k}|c_k|
\end{equation}

with the assumption that people with similar educational background should value the utilities in a similar way, we also tried to fit the utility parameters at the aggregated level, using the sum of the target functions at individual level as the target function. But the result was not promising.
{Both two estimation methods use the minimize function from Scipy.optimize library, with the Nelder Mead algorithm. Tolerance were set to 1e-6 and maximum iterations of 3000. The starting point was determined using a simple local grid search.}}

\subsubsection{Evaluation.}
In order to validate the model, we utilize K-fold cross-validation, where K=6 for both experiments 1 and 2. We have also implemented Random Forest, XGboost, and feed-forward neural network for the prediction problem and compared the performance of our model with them. For each of these machine learning models, we also used 6-fold cross-validation. The number of trees in the Random Forest was 100. The artificial neural network is a fully connected feed-forward neural network with one hidden layer of 10 neurons, using relu activation function and solved using Adam optimization algorithm. The models and hyperparameters are tuned using Dataiku AutoML framework with sk-learn as back-end machine learning library. The information available for every model are the same as the QDT models.

\begin{figure}[!b]
\vspace{-2mm}
~~~~~\includegraphics[width=0.9\linewidth]{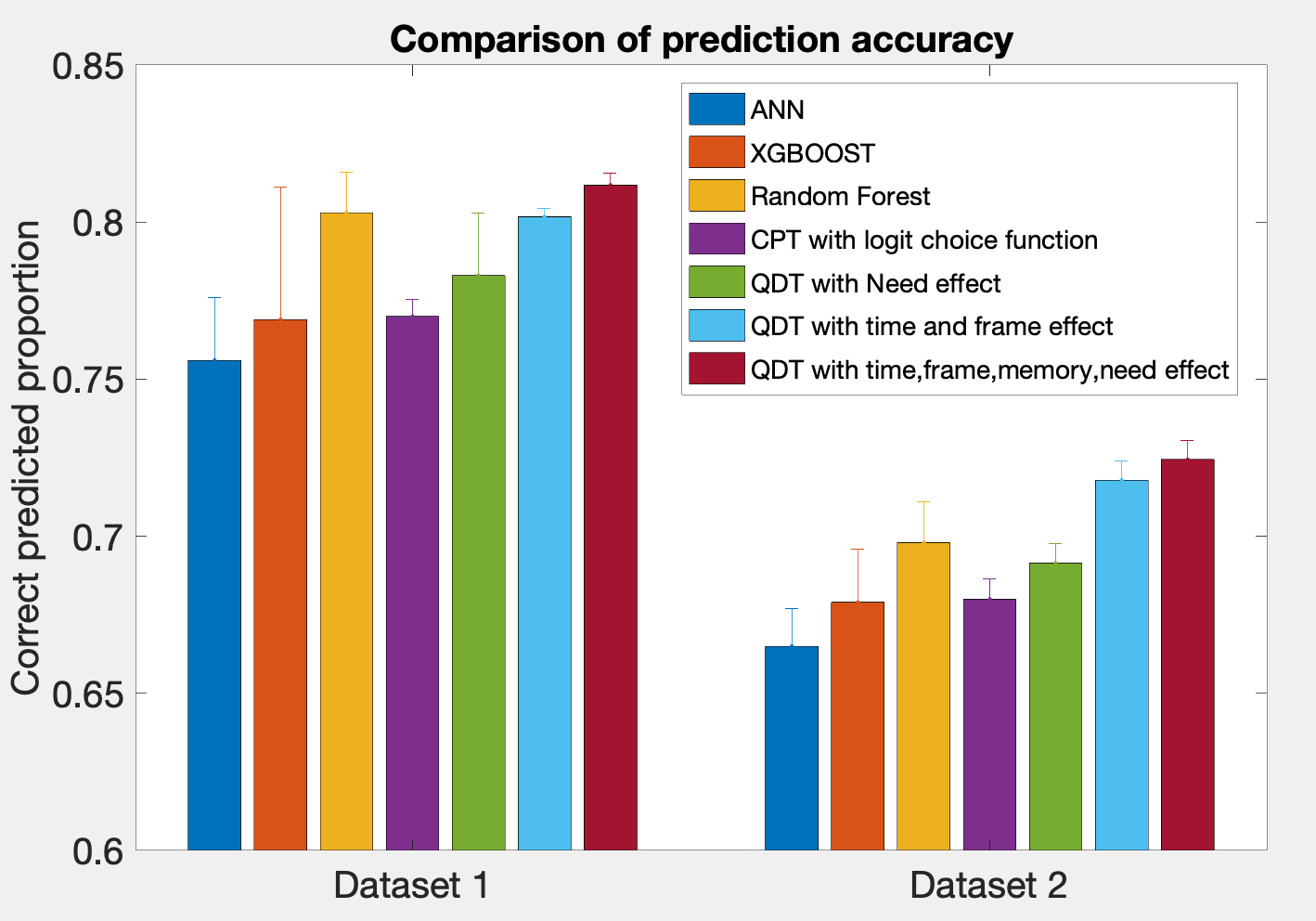}
\vspace{-2mm}
\caption{Accuracy comparison for different models} 
\label{accuracy}
\end{figure}

\section{Results}
{We measure the performance of the model with two measures, prediction accuracy and the ability to predict the option probabilities.}

\begin{figure*}[t]
\centering
\begin{subfigure}[t]{0.48\textwidth}
\centering
\includegraphics[width=0.75\linewidth]{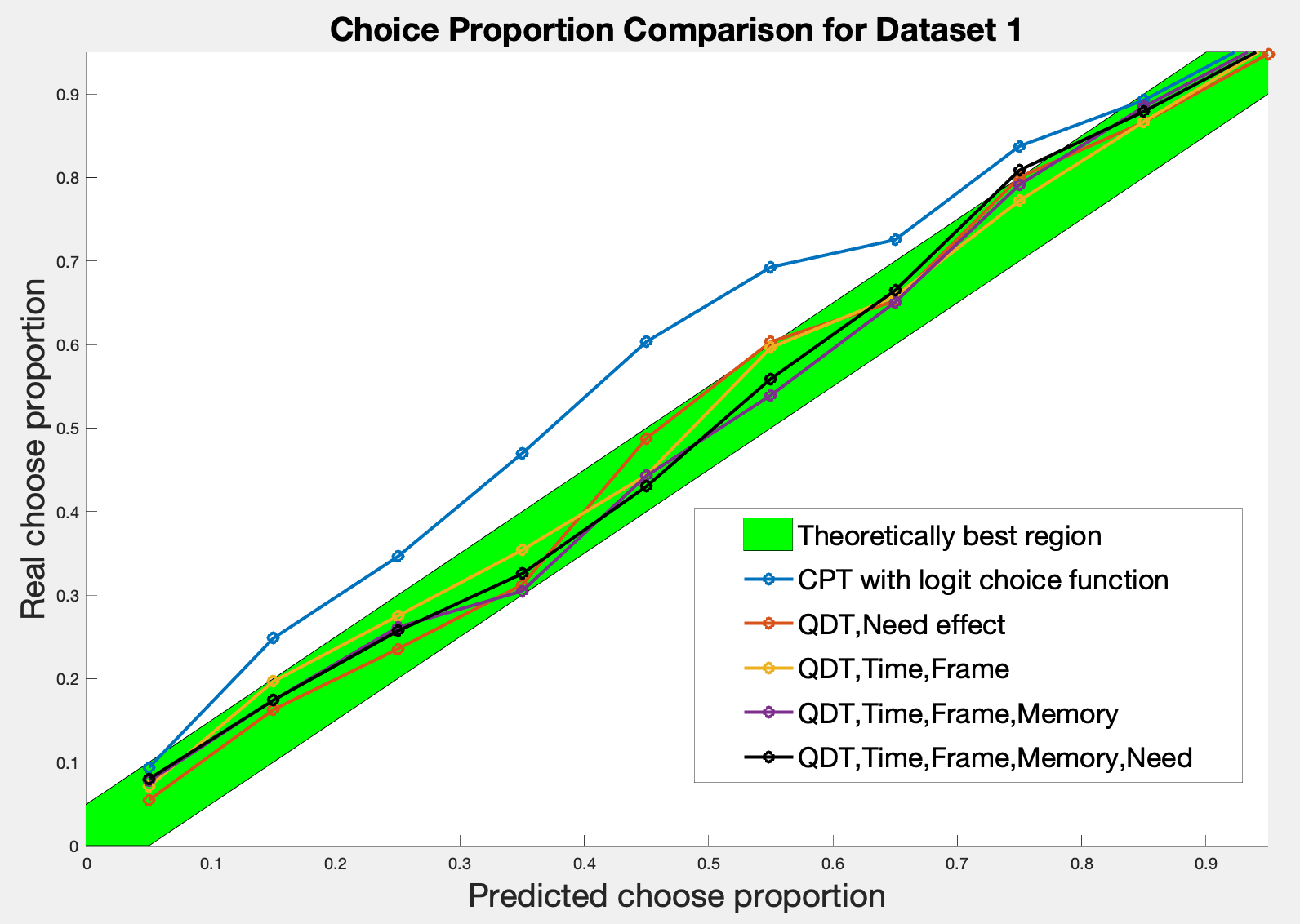}
\subcaption{Dataset 1}
\end{subfigure} ~~~~~~
\begin{subfigure}[t]{0.48\textwidth}
\centering
\includegraphics[width=0.75\linewidth]{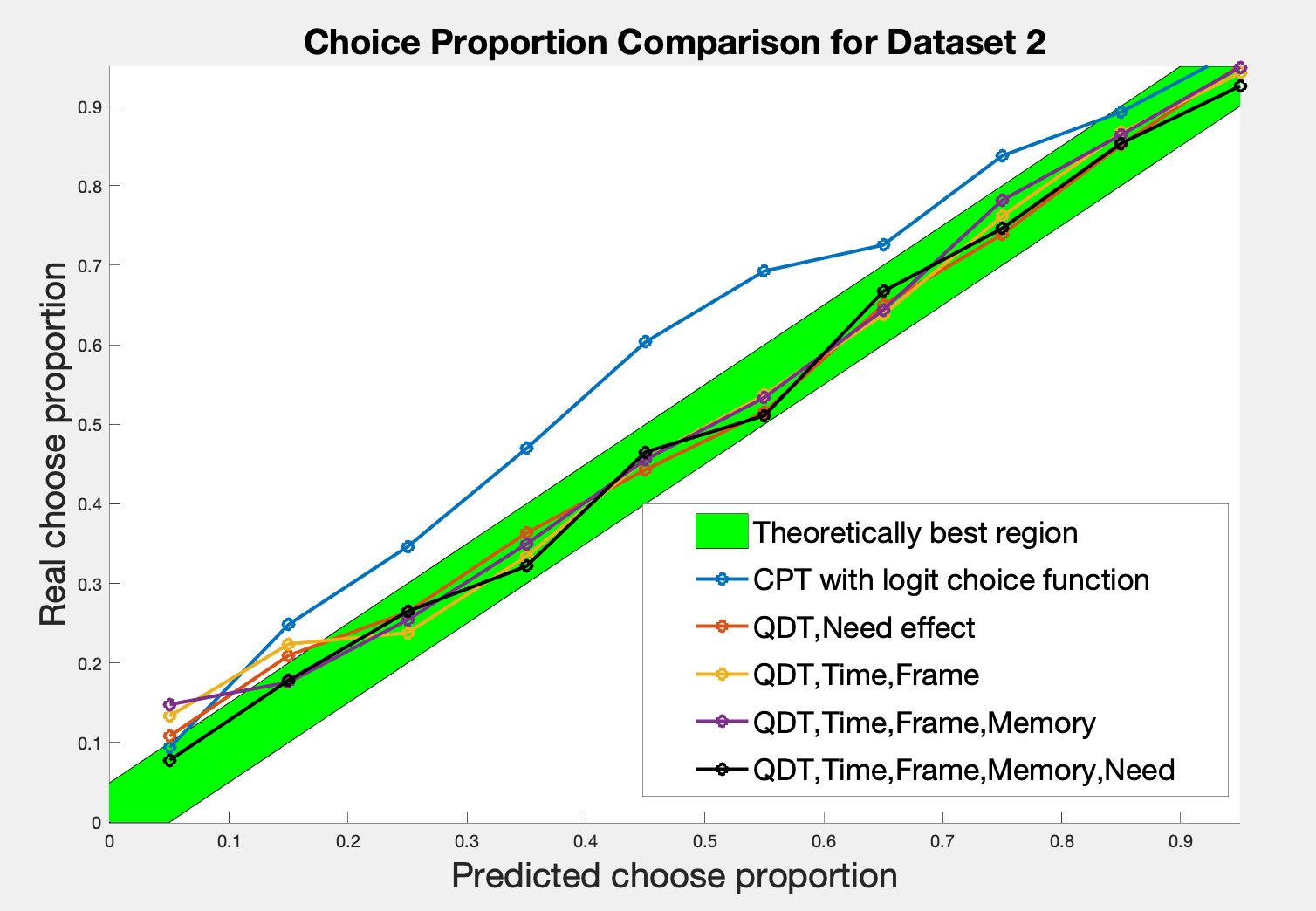}
\subcaption{Dataset 2}
\end{subfigure}
\caption{The ability of predicting the probabilities of picking each option. Game trials are grouped into ten groups according to QDT calculated option probability for the gamble option. For each group, the proportion of game trials that the gamble option has been chose in the empirical data is plotted. The ideal result is that all ten points lie in the green interval. The vertical width of the green interval is 0.1 as each group represents a probability interval of length 0.1. } 
\label{comp}
\end{figure*}

\subsubsection{Accuracy.}
{Figure \ref{accuracy} represents the prediction accuracy of different models. For the QDT models, We choose the option with higher calculated option probability as the prediction result. The prediction accuracy is calculated as the proportion of correctly predicted game trials in the test dataset.  Figure \ref{accuracy} demonstrates the accuracy comparison between the CPT model, three data-driven models, and three QDT models with different attraction factor functions. All the models perform better on the first experiment than the second one. There could be two reasons, data size difference and repeated game trials. Dataset 1 has \hk{more} 
data points per subject than Dataset 2 (960 versus 624), which means there is more information available for each subject in Dataset 1. There are also repeated game trials in Dataset 1, that are game trials with exactly the same settings. Although the player does not always choose the same option for the repeated game trials, this should still provide extra information to the models, especially when the repeated trials exist in both training and testing datasets. 
 Among three data-driven models, ANN performs the worst and Random Forest achieves the highest accuracy for both two datasets. The biggest issue for ANN could be the small data size, there are just not enough data points for it to perform well. Random forest with 100 decision trees performs surprisingly well on Dataset 1, achieved an average accuracy of $80.03\%$. On the other hand, all three QDT models perform better than the CPT model. QDT using all four attraction components performs the best among all the models, surpasses Random Forest's performance in both datasets.

We treat the time effect and frame effect attraction factor component together as a single attraction component because the time effect component only serves as a multiplier for the frame effect component. Tables \ref{accuracy_data1} and \ref{accuracy_data2} demonstrate the accuracy performance for different attraction factor components and their combinations. All three attraction factor components, when working on its own, increases the accuracy in comparison with the logit-CPT.

Time plus frame effect attraction factor components perform the best when using alone, while memory and need effect components' contribution are similar. When combing the memory component with Time+Frame effect components, the performance increase is not significant for both two datasets. After adding $q_{need}$, the QDT model achieves the highest prediction accuracy for both datasets. 

\begin{figure}[!b]
\centering
\includegraphics[width=0.75\linewidth]{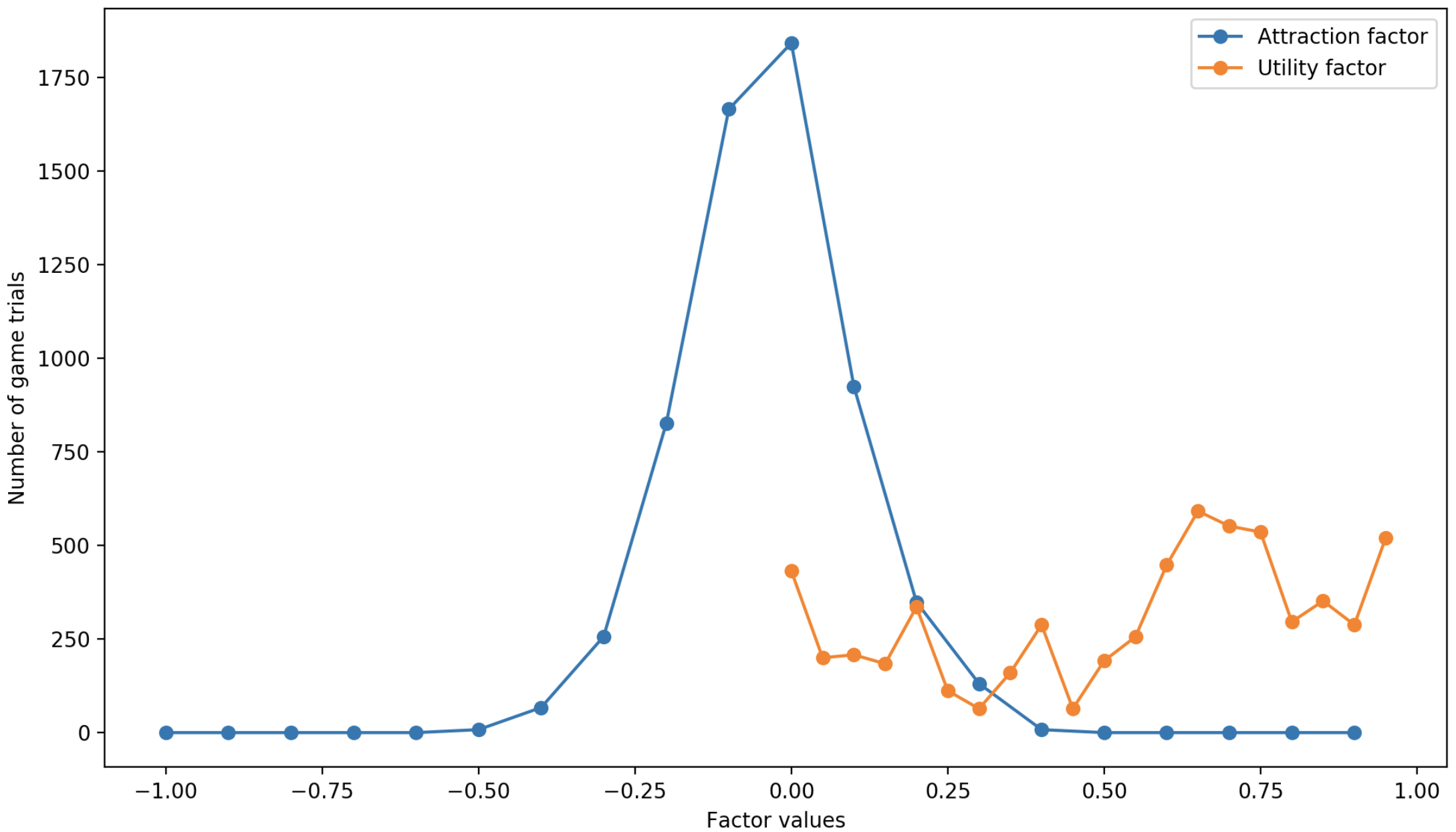}
\caption{The distribution of utility factor values and attraction factor values are shown in the same plot.} 
\label{FactorValue}
\end{figure}

We would like to conclude that combining different attraction factor components is useful, but the performance improvement seems not to be additive. Combing two useful attraction factor components does not necessarily improve the prediction accuracy. 

If the choices are probabilistic, there will also be a hard barrier preventing the accuracy from being further increased. This is also mentioned by \cite{vincent2016calibration} and \cite{murphy2018hierarchical}. This could also be an explanation to the performance difference between the two datasets. Game trials in dataset 1 could be more biased or easier to decide than game trials in dataset 2. }

 \subsubsection{Option probabilities.}
To evaluate the ability of calculating the option probabilities, we group all the game trials into ten groups according to the predicted gamble option probabilities. The first group contains games which the model predicts the decision-maker will have 0-10\% chance of picking the gamble option, the second group contains games with 10-20\% option probability, and so on. We then count the empirical proportion of choosing the gamble option within each group and check if they actually fall into the calculated probability interval. For example, for the first group, the ideal case would be only $0\%-10\%$ of games trials received gamble option as response in the reality. If we assume a uniform distribution of predicted option probabilities within each interval, the most optimal empirical choose proportion would be the midpoint of each interval, that is, $5\%$ for the first interval, $15\%$ for the second.

Figure \ref{comp} shows the empirical choice proportion against the predicted option probability. We observe that nearly all the QDT models predict the option probability well, where the empirical statistics always stay within the acceptable interval. QDT using all four attraction factor components performs as best as the empirical statistics are closer to the middle point of each interval. On the other hand, CPT with logit choice function seems to overestimate the probabilities on both two datasets.

\subsubsection{Effect of including catch trials.}
Catch trials are game trials with unfair options. For example, a gamble biased catch trial could have a gamble option of keeping 100 points with 70$\%$ chance and a sure option of keeping 40 points. 
We found out that including the catch trials in the training data set boosts the performance of both CPT and QDT models. 

\begin{figure}[!b]
\centering
\includegraphics[width=0.75\linewidth]{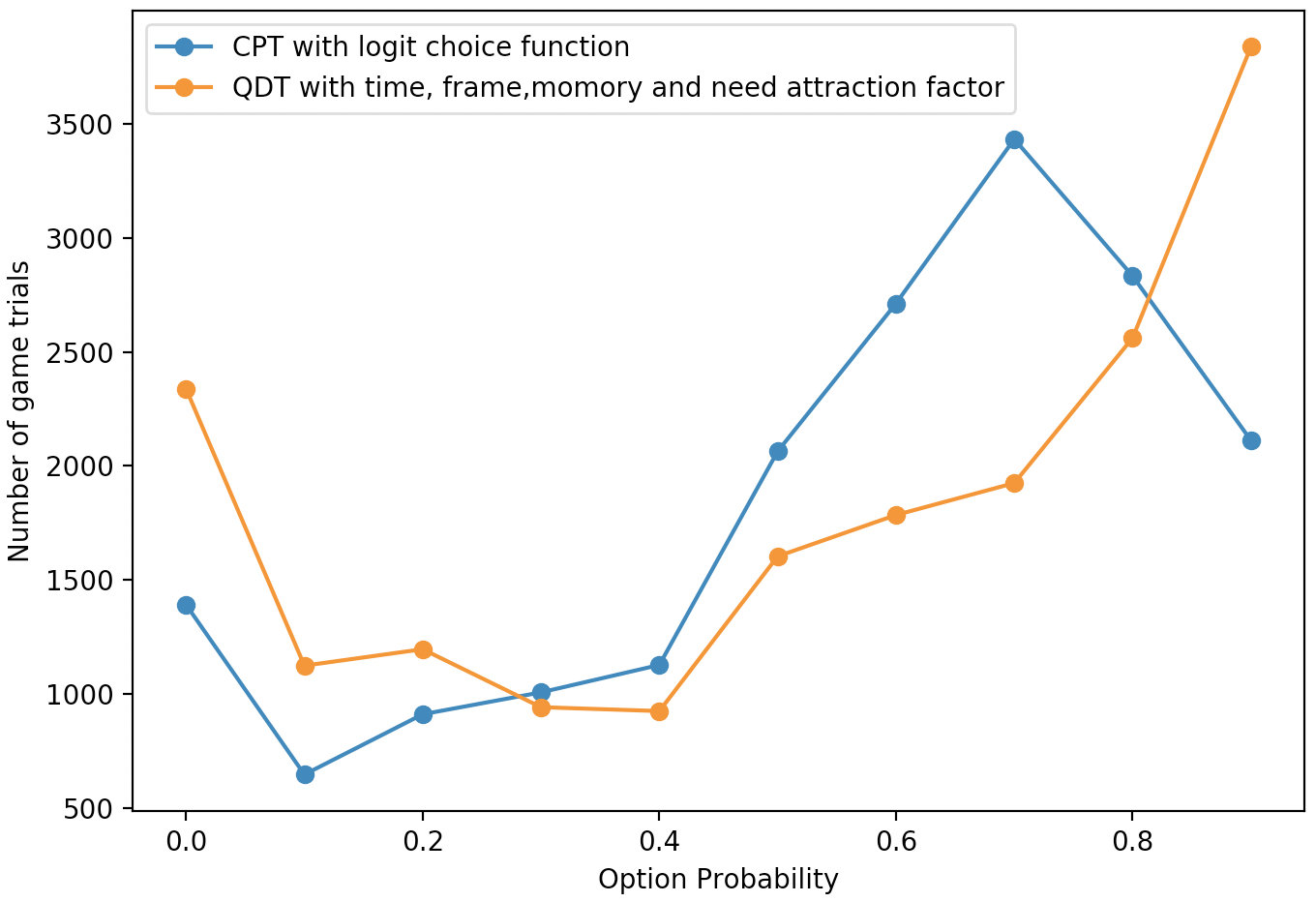}
\caption{Number of game trials in each probability interval.} 
\label{optionprob}
\end{figure}

\begin{figure*}[!t]
\centering
\begin{subfigure}[t]{0.48\textwidth}
~~~~\includegraphics[width=0.9\linewidth]{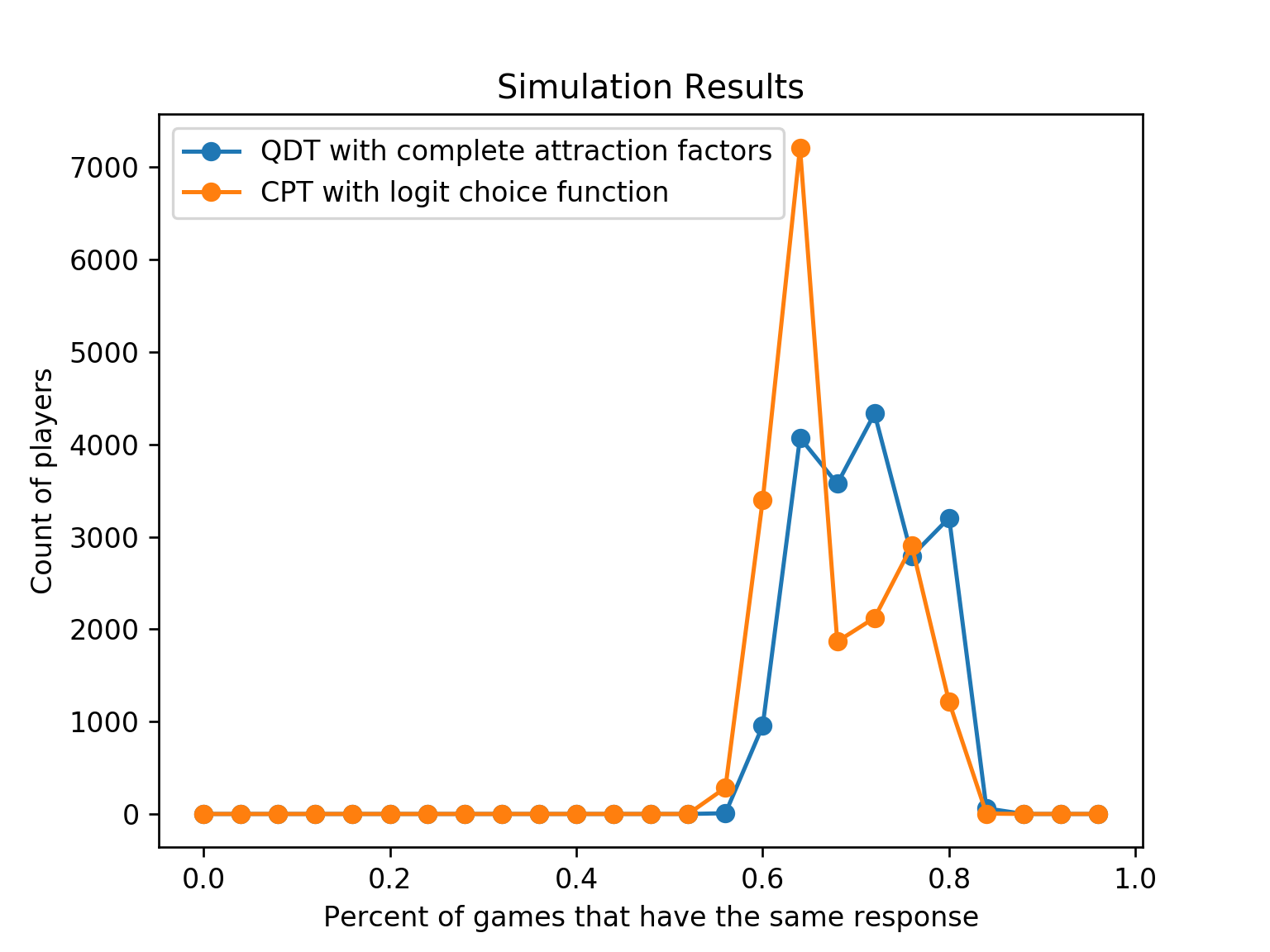}
\subcaption{Dataset 1}
\end{subfigure} ~~~~~~
\begin{subfigure}[t]{0.48\textwidth}
~~~~\includegraphics[width=0.9\linewidth]{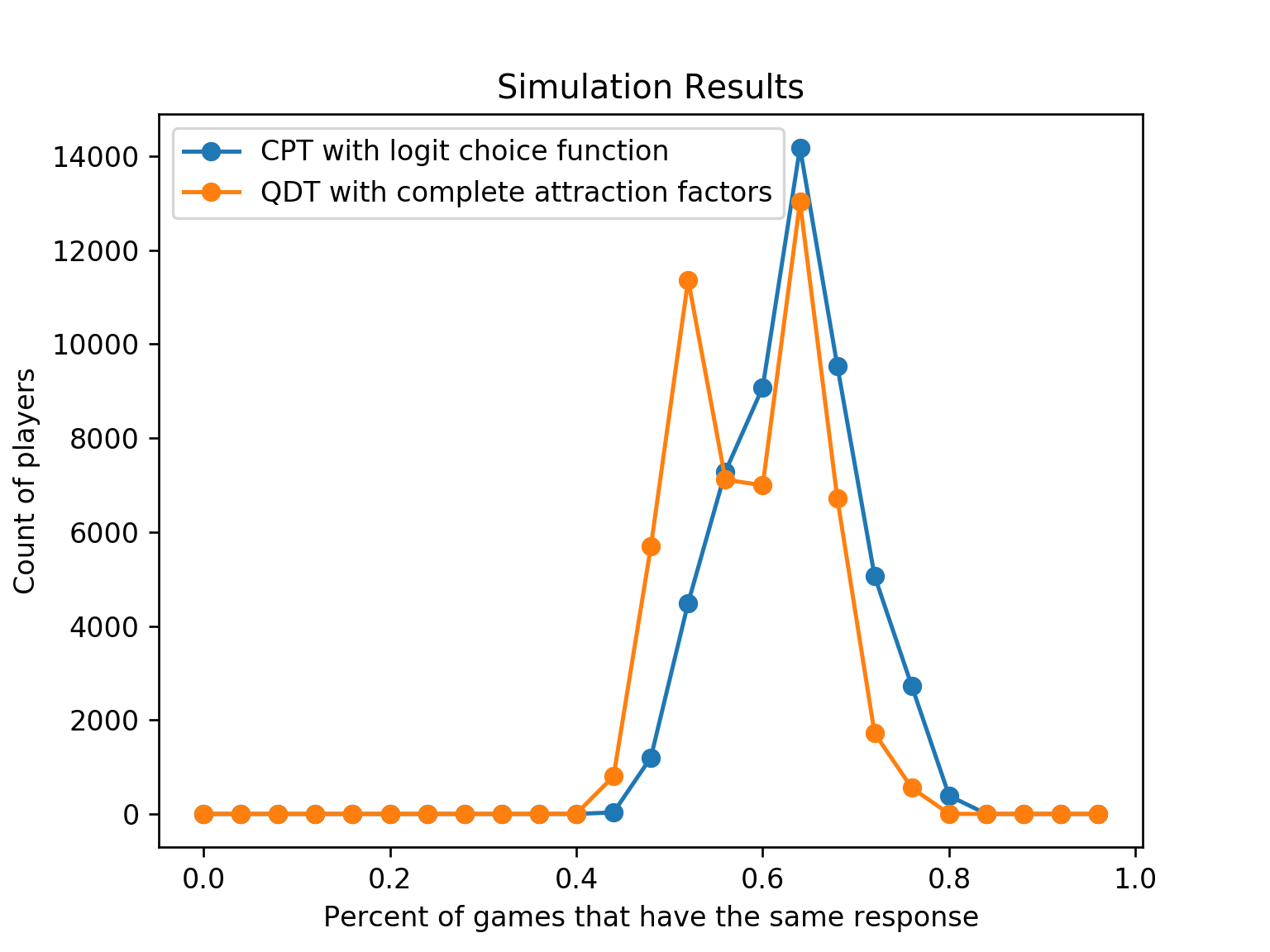}
\subcaption{Dataset 2}
\end{subfigure}
\caption{Simulation results for 1000 simulations. For each subject in each simulation, we calculate the proportion of game trials that has the same response as the empirical dataset. The distribution of the response similarity is plotted. } 
\label{simulation}
\end{figure*}

For Dataset 
2, we observe a performance boost of approximately $2\%$ for both CPT and QDT models. The catch trials has the potential to capture some 
influential subconscious effects. Because there exists a clearly better option, any response that chooses the "wrong" option could be caused by 
irrational aspects of the decision maker's state of mind, such as great risk aversion under extreme time constraint or willing of avoiding uncertainty when the participant's score is very close to the minimum needed score. The amount of such "abnormal" responses in the catch trial reflect how much do the subconscious effects impact the participant, thus including the catch trials brings performance boost. 

\subsubsection{Factor values and probability distribution.}
Figure \ref{FactorValue} shows the distribution of utility factor values f($\pi_{gamble}$) and attraction factor values q($\pi_{gamble}$) on an aggregated level for dataset 1. The utility factor f($\pi$) is defined between 0 and 1; thus, the orange line only exists in that interval. The utility factor values are distributed in a relatively uniform way, without a significant peak. There are comparably more game trials that have a utility factor value greater than 0.6, which indicates many participants in Dataset 1 are more inclined towards the gamble option without considering the impact of subconscious effects, even though every game trials are fair. On the other hand, the attraction factor values are distributed in a Gaussian-like distribution with a mean at zero. There are nearly no game trial and participant that gives a very extreme attraction factor values. The reason could be that the absolute value of the attraction factor is bounded by the minimum value between two utility factors, which means the absolute value of the attraction factor would never exceed 0.5. Besides, considering there are significantly many game trials that have polarized utility factor values, either a small gamble option utility factor or a small sure option utility factor, the small amplitudes of attraction factors are not hard to understand. There are nearly equally many game trials on the positive side and negative side, one primary reason could be the intrinsic symmetric design of the experiments. E.g.~for the framing effect; half of the game trials have a sure option described in a losing frame, while another half with a gain frame description. Therefore the values of the frame effect components will also be distributed symmetrically. 

Figure \ref{optionprob} demonstrates how the 
probability of the gamble option is distributed. The probabilities predicted by QDT are more polarized than the ones 
by CPT; there are many more game trials with very high or very low values of option probability. In contrast, most of CPT's option probabilities concentrate at a relatively average interval, between 0.5 and 0.75.  The attraction factors seem to add another level of preference on top of the classical utility preference, making one option much more preferable than the other option.

\subsubsection{Simulations using QDT calculated 
option probabilities.}
QDT is a probabilistic framework. The decision-maker does not make deterministic decisions. Therefore we use the option probabilities calculated by QDT and CPT to simulate the experiments and compared them with empirical data. For both Dataset 1 and 2, 1000 simulations are done for each participant. We then analyzed the similarity between the simulated players' answers with real participants' answers by inspecting the proportion of game trials that the simulated player answered the same as the real participant. We counted the number of simulated players based on the response similarity and Figure \ref{simulation} show the results. For both datasets 1 and 2, the QDT model performs better than the CPT model; there are more simulated players with higher response similarity. We also observe that the distributions are multimodal. By the central limit theorem, we are expecting the distribution for a single participant to be normally distributed. When aggregating the result for all participants, the aggregated distribution becomes a mixture of normal distributions. The modes of the distributions indicate there is a group of participants that have been modeled similarly well by the model. For Dataset 1, the CPT model simulates one group of participants well, with a peak at 0.75, but has a much lower response similarity for the rest of the participants as we see the other peak at 0.625. On the other hand, all three modes of QDT model's distribution are relatively higher. This indicates that the QDT models predict more polarized option probabilities, which also agrees with the result shown in Figure \ref{optionprob}. The situation is the same for Dataset 2, both CPT and QDT have a peak at 0.625, but CPT has another peak at lower response similarity of 0.5, which means there is a group of participants that the CPT model predicts very fair option probabilities between the gamble option and sure option. The added attraction factors capture the subconscious effects of these participants and thus, the option probabilities become more polarized, indicating one option is more preferable than the other. Therefore the QDT models always have a higher response similarity.

\section{Conclusions}
We present \hk{a method for} 
modeling human decision-making using \hk{QDT}, 
using \hk{several} different attraction factor components to capture \hk{a range of irrational and subjective aspects of the decision-making. Experiments show that this modeling of the decision-maker's state of mind enables a more accurate prediction of individual decisions than with classical CPT, which only models limited cognitive aspects. Moreover, the proposed model outperforms a range of data-driven methods.}

The parameters are now tuned individually. To allow faster adaptation to new individuals, a good option is hierarchical parameter estimation \cite{murphy2018hierarchical} with training first on the group level and then fine-tuning for each individual. 
Another future direction could be formulating more generally applicable attraction factor components that can be used in a variety of decision problems. Another interesting direction is to combine to model with deep neural networks. One major problem in applying deep neural networks in predicting human decisions is the lack of data. Our QDT model could be served as a cognitive model that can be used to generate massive human decision data. Approximate the attraction factor using neural networks could also be a direction worth exploring  




\bibliography{ConfPaper.bib}
\appendix
\section{Appendix}
This appendix is borrowed from \cite{vincent2016calibration}, developed by Yukalov and Sornette in a series of articles.\cite{yukalov2008quantum}\cite{yukalov2009processing}\cite{yukalov2010mathematical}. Quantum decision theory (QDT) has recently been introduced as an alternative formulation to existing theories. It is based on two essential ideas: (i) an intrinsic probabilistic nature of decision making and (ii) a generalisation of probabilities using the mathematics of Hilbert spaces that naturally accounts for entanglement between choices.
\subsection{Mathematical structure of QDT}
\textbf{Definition.1(Action Ring)} \\The action ring A = $\{A_n:n=1,2,...N\}$ is the set of intended actions, endowed with two binary operations:\\
-The reversible and associative addition.\\
-The non-distributive and non-commutative multiplication, which possesses a zero element called empty action.\\
The interpretation of the sum A + B is that A or B is intended to occur. The product AB means that A and B will both occur. The zero element is the impossible action, so AB = BA = 0 means that the actions A and B cannot occur together: they are disjoint.\\
\textbf{Definition.2(Composite action and action modes)}\\When an action $A_n$ can be represented as an union (i.e. is the sum of several actions), it is referred to as composite. Otherwise it is simple.\\
The particular ways $A_{jn}$ of realizing a composite action $A_n$ are called the action modes and are disjoint simple elements:
$$A_{n}=\bigcup_{j}^{M_{n}} A_{j n} \quad M_{n}>1$$
\textbf{Definition.3(Elementary prospects)}\\ An elementary prospect $e_{\alpha}$ is an intersection of separate action modes,
$$e_{\alpha}=\bigcap_{n} A_{\alpha n}$$
where the $A_{\alpha n}$ are action modes such that $e_{\alpha}e_{\beta}$ = 0 if $\alpha \not = \beta$.\\
\textbf{Definition.4(Action prospect)}\\ A prospect $\pi_n$ is an intersection of intended actions, each of which can be simple (represented by a single action mode) or composite.\\
$$\pi_{n}=\bigcap_{j} A_{n_{j}}$$
To each action mode, we associate a mode state $\left|A_{j n}\right\rangle$ and its hermitian conjugate $\left\langle A_{j n}\right|$ . Action modes are assumed to be orthogonal and normalized 
to one, so that $\left\langle A_{j n} \mid A_{k n}\right\rangle=\delta_{j k}$ . This allows us to deﬁne orthonornal basic states for the elementary prospects:
$$|e_{\alpha}\rangle =|A_{\alpha 1} \ldots A_{\alpha N} \rangle$$
$$\langle e_{\alpha} \mid e_{\beta}\rangle=\prod_{n} \delta_{\alpha_{n}} \delta_{\beta_{n}}=\delta_{\alpha \beta}$$

\textbf{Definition.5(Mind space and prospect state)}\\ The mind space is the Hilbert space
$$\mathcal{M}=Span\left\{\left|e_{\alpha}\right\rangle\right\}$$
For each prospect $\pi_n$ , there corresponds a prospect state $\left|\pi_{n}\right\rangle \in \mathcal{M}$
$$\left|\pi_{n}\right\rangle=\sum_{\alpha} a_{\alpha}\left|e_{\alpha}\right\rangle$$
\textbf{Definition.6(Strategic state of mind)}\\ The strategic state is a normalized ﬁxed state of the mind space M describing a decision maker at a given time:
$$|\psi\rangle=\sum_{\alpha} c_{\alpha}\left|e_{\alpha}\right\rangle$$
The strategic state characterizes a particular decision maker at a given time, it includes his/her personal attributes and is related to the information available to the decision maker.
\subsection{Prospect probabilities}
In the context of quantum decision theory, the preferences of a decision maker depend on his/her state of mind and on the available prospects. Those preferences are expressed through prospect operators.\\
\textbf{Definition.7(Prospect Operator)}\\ For each prospect $\pi_n$ , we deﬁne the prospect operator\\
$$\hat{P}\left(\pi_{n}\right)=\left|\pi_{n}\right\rangle\left\langle\pi_{n}\right|$$
By this deﬁnition, the prospect operator is self-adjoint. Its average over the state of mind deﬁnes the prospect probability $p(\pi_n)$ :
$$p\left(\pi_{n}\right)=\left\langle\psi\left|\hat{P}\left(\pi_{n}\right)\right| \psi\right\rangle$$
The decision maker is more likely to choose the prospect with the highest prospect probability. The probabilities should correspond to the frequency with which the prospect would be chosen if the choice could be made several times in a same state of mind.

By deﬁnition .5 and .6, we can distinguish two terms in the expression of $p(\pi_n)$: a utility factor f($\pi_n$) and an attraction factor $q(\pi_n)$ :
$$p\left(\pi_{n}\right)=f\left(\pi_{n}\right)+q\left(\pi_{n}\right)$$
$$f\left(\pi_{n}\right)=\sum_{\alpha}\left|c_{\alpha}^{*} a_{\alpha}\right|^{2}$$
$$q\left(\pi_{n}\right)=\sum_{\alpha \neq \beta} c_{\alpha}^{*} a_{\alpha} a_{\beta}^{*} c_{\beta}$$
Within the framework of quantum decision theory, the utility and attraction terms are subjected to additional constraints:
$$f\left(\pi_{n}\right) \in[0,1], \sum f\left(\pi_{n}\right)=1$$
$$q\left(\pi_{n}\right) \in[-1,1],\sum q\left(\pi_{n}\right)=0$$

\balance
\end{document}